%% file: root.tex
\documentclass[letterpaper, 10 pt, conference]{ieeeconf}  % Comment this line out if you need a4paper

\IEEEoverridecommandlockouts                              % This command is only needed if 
\usepackage{graphics} % for pdf, bitmapped graphics files
\usepackage{epsfig} % for postscript graphics files
\usepackage{mathptmx} % assumes new font selection scheme installed
\usepackage{times} % assumes new font selection scheme installed
\usepackage{amsmath} % assumes amsmath package installed
\usepackage{amssymb}  % assumes amsmath package installed

\usepackage[usenames,dvipsnames]{xcolor}
\usepackage{color, colortbl}
\definecolor{Gray}{gray}{0.94}

\usepackage{hyperref}
\hypersetup{
    colorlinks=false,     % Set this to false to have boxed links instead of colored text
    linkcolor=blue,      % Color of internal links
    citecolor=green,     % Color of citation links
    urlcolor=blue,       % Color of external links
}
\usepackage{booktabs}
\usepackage{multirow}
\usepackage{caption}
\usepackage{subcaption}
\title{\LARGE \bf
C2FDrone: Coarse-to-Fine Drone-to-Drone Detection using Vision Transformer Networks
}

\author{ Sairam VC Rebbapragada$^{1 \dagger}$, Pranoy Panda$^{2}$ and Vineeth N Balasubramanian$^{1}$  % <-this % stops a space
\thanks{$^{1}$ is with Machine Learning and Vision group, IIT Hyderabad, India
        {\tt\small people.iith.ac.in/vineethnb/students.html}}%
\thanks{$^{2}$ is with Fujitsu AI Research India, but this work was done while affiliated to IIT Hyderabad.}%
\thanks{$^\dagger$ corresponding author: ai20resch13001@iith.ac.in}
}

\begin{document}

\maketitle
\thispagestyle{empty}
\pagestyle{empty}

%%%%%%%%%%%%%%%%%%%%%%%%%%%%%%%%%%%%%%%%%%%%%%%%%%%%%%%%%%%%%%%%%%%%%%%%%%%%%%%%
\begin{abstract}

% A vision-based drone-to-drone detection system offers a cost-effective solution for a range of applications, including collision avoidance, countering hostile drones, and enhancing search-and-rescue operations. However, drone-to-drone detection presents a more intricate set of challenges compared to regular object detection. These challenges encompass the need to detect extremely small-sized objects, contend with strong distortion, handle severe occlusion, operate in uncontrolled environments, and execute real-time processing. While current methods attempt to address these issues by integrating multi-scale feature fusion and temporal information, we propose that these techniques may not be sufficiently equipped to handle extreme blur and minuscule objects. Instead, we put forth a novel coarse-to-fine detection strategy based on vision transformers to achieve precise drone detection. We assess the effectiveness of our approach through a series of comprehensive experiments conducted on three challenging drone-to-drone detection datasets. Our results demonstrate notable improvements, with F1 score enhancements of $7\%$, $3\%$, and $1\%$ on the FL-Drones, AOT, and NPS-Drones datasets, respectively. Furthermore, we showcase its real-time processing capability by deploying our model on an edge-computing device. We will make our code repository publicly available.
A vision-based drone-to-drone detection system is crucial for various applications like collision avoidance, countering hostile drones, and search-and-rescue operations. However, detecting drones presents unique challenges, including small object sizes, distortion, occlusion, and real-time processing requirements. Current methods integrating multi-scale feature fusion and temporal information have limitations in handling extreme blur and minuscule objects. To address this, we propose a novel coarse-to-fine detection strategy based on vision transformers. We evaluate our approach on three challenging drone-to-drone detection datasets, achieving F1 score enhancements of $7\%$, $3\%$, and $1\%$ on the FL-Drones, AOT, and NPS-Drones datasets, respectively. Additionally, we demonstrate real-time processing capabilities by deploying our model on an edge-computing device. Our code will be made publicly available.

\end{abstract}

%%%%%%%%%%%%%%%%%%%%%%%%%%%%%%%%%%%%%%%%%%%%%%%%%%%%%%%%%%%%%%%%%%%%%%%%%%%%%%%%
\section{INTRODUCTION} \label{sec:introduction}

In recent years, drones have demonstrated remarkable versatility in various fields such as agriculture \cite{10176980, 10041534}, military operations, search-and-rescue missions \cite{10118136}, firefighting \cite{9156137}, aerial photography, and essential deliveries \cite{9074487}. This increasing demand for drones has prompted extensive research into enhancing their vision capabilities, particularly object detection \cite{han2021redet, cao2021visdrone, sairam2023aruba, Hua_2023_CVPR}. Along with detecting other objects on the ground, it is equally important for drones to detect each other in the air. This capability helps avoid drone collisions, counter hostile drones, and facilitates drones to collaborate and cover larger areas during search-and-rescue operations. While research on drone-based ground object detection has been well-studied, drone-to-drone detection remains relatively less explored. %\cite{ashraf2021dogfight, sangam2023transvisdrone}.

\begin{figure}[!th]
    \centering
    \includegraphics[width=\linewidth]{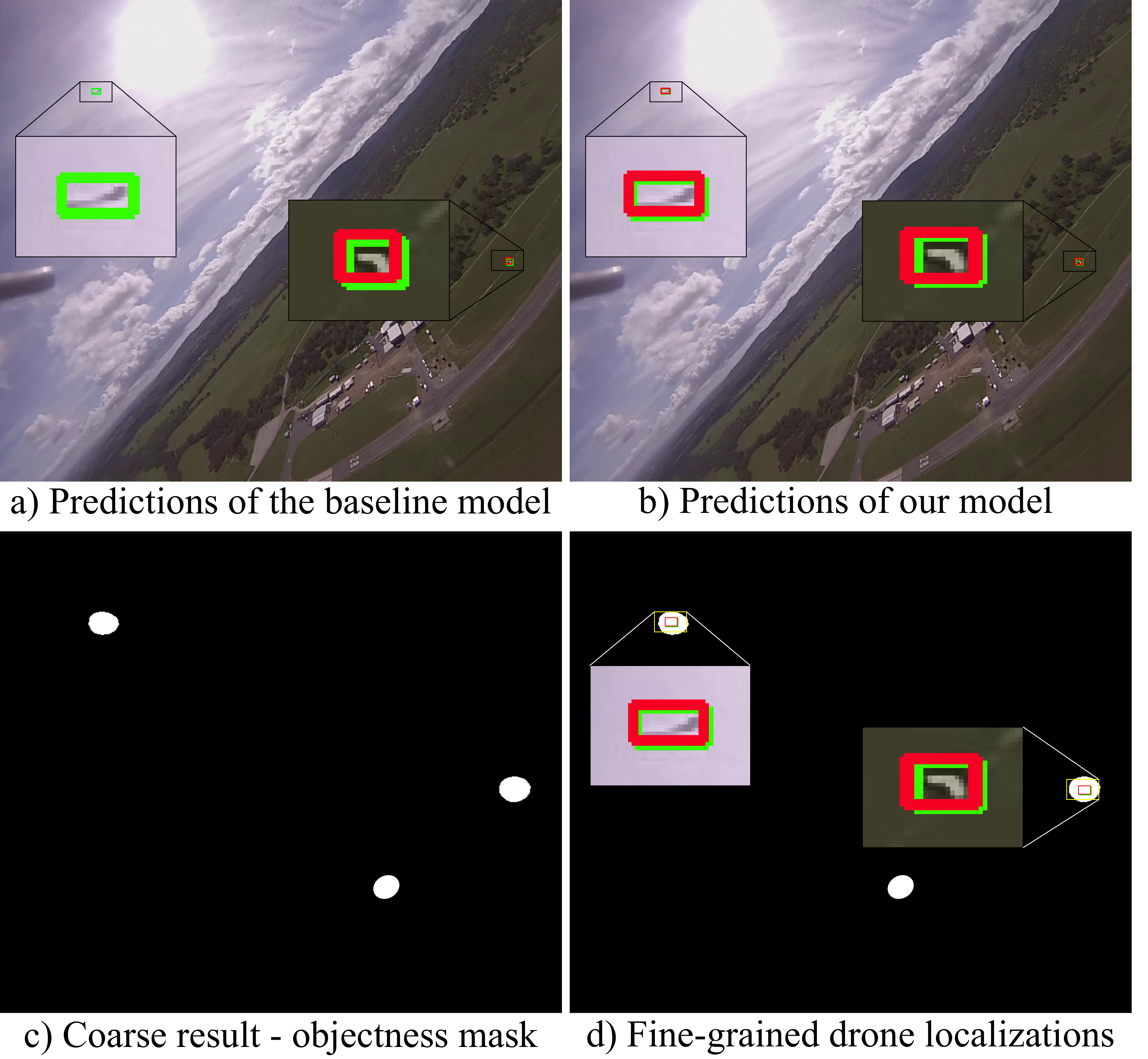}
    \vspace{-5mm}
    \caption{A challenging frame from NPS Drones dataset \cite{li2016multi}. \textcolor{Green}{Green} boxes - ground truth, \textcolor{red}{Red} boxes - model predictions a) Traditional methods uniformly scan the entire frame for drones, leading to wasted effort and missed detections in complex scenarios b) Our method precisely localizes drones using a coarse-to-fine detection approach c)  Coarse level narrows down the search space by generating an objectness mask d) Fine-grained level focuses on the refined search space, enhancing drone detection.
    % a) Traditional methods provide equal attention to the entire frame, missing drones that are blended with the background, or heavily distorted. b) Our coarse-to-fine approach localizes all the drones by systematically narrowing the search space. c)  Coarse level reduces noise in the feature space and generates an objectness mask highlighting likely object regions. d) Fine-grained level focuses on the reduced search space, enhancing drone detection.
    }
    \label{figure_1}
\vspace{-5mm}
\end{figure}

\vspace{0.5mm}
Drone-to-drone detection presents a more complex set of challenges when compared to regular object detection. These challenges encompass the detection of extremely small-sized objects, dealing with strong distortion, handling severe occlusion, operating in uncontrolled environments, and the requirement for real-time processing. In drone-to-drone scenarios, the captured videos are likely to contain heavy noise and distortion because both source and target drones are in constant motion, and the cameras onboard may not always be high-resolution \cite{rozantsev2016detecting}. 
% When we pass these noisy and distorted video frames through the convolutional neural network (CNN) backbones or other feature extraction architectures commonly used for object detection, downsampling operations like pooling or strided convolutions are typically applied. These operations are used to reduce the spatial resolution of the input image while increasing the receptive field of the network. However, in the presence of heavy noise and distortion, downsampling can exacerbate the problem. 
When using convolutional neural networks (CNNs) or similar feature extraction architectures for object detection, downsampling operations like pooling or strided convolutions are commonly applied. However, in the presence of heavy noise and distortion, downsampling might potentially exacerbate the problem. Thus, it is crucial to devise effective strategies for noise reduction in the extracted features. Additionally, when operations like max-pooling are employed, essential local information is lost, which is detrimental to detecting small-sized objects. Recent approaches like \cite{ashraf2021dogfight} and \cite{sangam2023transvisdrone} have incorporated temporal information to address blurriness and occlusion issues, using a multi-resolution feature fusion approach to capture small objects. While effective to some extent, these methods may not be optimal for extreme distortion cases and scenarios where drones are tiny and blend into their backgrounds.

\vspace{0.5mm}
In this paper, we hypothesize that relying on simple multi-scale feature fusion and indiscriminately allocating equal attention to the entire frame is not sufficient for accurately localizing drones in real-world scenarios (Fig. \ref{figure_1}a). To substantiate this hypothesis, we present empirical findings (Section \ref{sec:ablation_studies}) and qualitative results (Section \ref{sec:qualitative_res}) as evidence and propose a novel coarse-to-fine detection strategy using Vision Transformer networks \cite{liu2021swin}, \cite{liu2022dabdetr} to systematically reduce the search space for drones and enhance the drone-to-drone detection performance (Fig. \ref{figure_1}b). At the coarse level (Fig. \ref{figure_1}c), we reduce the noise in the feature space and identify regions within the frame more likely to contain objects. Subsequently, at the fine-grained level (Fig. \ref{figure_1}d), we allocate increased attention to these identified regions. Our proposed method surpasses various competitive baselines on three benchmark datasets:  FL-Drones \cite{rozantsev2016detecting}, NPS-Drones \cite{li2016multi}, and AOT \cite{AOT}. 

\vspace{0.5mm}
To summarize, the major contributions of our work are:

\begin{itemize}
    \item We propose a novel coarse-to-fine detection strategy for localizing drones in drone-captured videos leveraging vision transformer networks and harnessing the untapped objectness information embedded in the image representations. Our method is designed to be end-to-end trainable and deliver real-time performance.
    % This approach is generic and can be seamlessly integrated into various object detection methods facing similar challenges, making it versatile and adaptable to different scenarios and applications.  
    % \item Our findings on mean feature map values in attention-based backbones offer valuable insights that we used for a coarse objectness mask; our insights herein can potentially be applied in other contexts, such as in open-world object detection where the model needs to discover unknown categories at runtime. %novel class discovparticularly in identifying regions of high object significance. For instance, this information can be utilized to classify unlabeled objects as 'unknown classes' in open-world object detection and classification tasks.
    % \item To the best of our knowledge, this is the first such approach to leverage the state-of-the-art DETR \cite{carion2020end, zhu2020deformable, liu2022dabdetr} family of models for drone-to-drone detection.
    \item We incorporate simple yet effective additions to the state-of-the-art DAB DETR \cite{liu2022dabdetr} model's design to achieve our coarse-to-fine detection objective. 
    \item We provide a comprehensive suite of experiments to validate the effectiveness of the proposed approach. We also carry out additional ablation studies and qualitative results to illustrate the usefulness of the proposed method in localizing drones. 
    % \item We perform a comprehensive suite of experiments on multiple datasets like FL-Drones, NPS-Drones and AOT to validate the effectiveness of the proposed approach. We also provide additional ablation studies and qualitative results to illustrate the usefulness of the proposed method in localizing drones. 
\end{itemize}

\section{RELATED WORK}

\subsection{Drone Detection}

The surge in drone usage has raised concerns about privacy and security threats. To address this, researchers have been developing effective drone detection methods. Some methods rely solely on non-visual sensor data like RF Sensors \cite{dressel2019hunting}. However, these methods are limited to drones with attached RF sensors. Another approach \cite{nava2022learning} involves self-supervised learning for quadrotor visual localization using its noise as a source of guidance. Additionally, point cloud data is utilized \cite{chen2021real} to segment voxels and navigate around obstacles, but this method requires expensive LiDAR sensors. Alternatively, LiDAR sensors on the ground are employed by \cite{dogru2022drone} to detect drones in the air. 

\vspace{0.5mm}
While using multiple sensors can enhance detection accuracy, it is beneficial to detect drones using cost-effective RGB cameras instead of costly radar systems to maintain the affordability and lightweight design of the drones. 
% Detecting drones in a drone-to-drone scenario presents more challenges than identifying drones captured by stationary ground cameras. 
\cite{yang2020autonomous} is an early work solely reliant on visual data. Their approach involves creating multiple spatio-temporal (s-t) tubes at various spatial resolutions. They employ two CNN models to achieve motion stabilization within each s-t tube. Drone detection is then carried out by classifying each s-t tube using a third CNN. Subsequently, \cite{ashraf2021dogfight} proposed a two-stage approach for drone-to-drone detection. In the first stage, they focus on spatial cues using CNNs and attention mechanisms. In the second stage, they leverage spatio-temporal information to reduce false positives and detect missed drones from the first stage. However, these approaches, \cite{yang2020autonomous} and \cite{ashraf2021dogfight}, suffer from being two-staged, computationally expensive, and impractical for deployment. Recently, \cite{sangam2023transvisdrone} proposed a real-time end-to-end methodology for drone-to-drone detection. It uses CSPDarkNet 53 \cite{wang2020cspnet} for extracting spatial features from a video clip and employs Video Swin Transformer \cite{liu2022video} to exploit video temporal information. However, both \cite{ashraf2021dogfight} and \cite{sangam2023transvisdrone} use a simple multi-scale feature fusion for detecting small-sized objects. This approach may not yield optimal results for extreme cases of distortion, camouflage, and tiny objects in real-world drone-to-drone detection scenarios as multi-scale feature extraction uses downsampling which might potentially enhance the inherent noise in the images, and when operations like max-pooling are utilized, local information in the features is lost which is important for detecting tiny objects. 

\subsection{DETR Models for Object Detection}

DETR, as presented in \cite{carion2020end}, provided a new perspective to object detection by offering an end-to-end trainable system that eliminates the need for handcrafted components such as non-maximum suppression and anchor generation. This innovative approach employs a transformer-based architecture to directly predict object class labels and bounding box coordinates for all objects in a single pass. Key components of DETR encompass positional encoding, the encoder, and the decoder. The encoder employs self-attention mechanisms to process image features and capture contextual information, while the decoder utilizes queries to attend to encoded features and make predictions. To address DETR's slow convergence, several variants have been introduced, including Deformable-DETR \cite{zhu2020deformable}, Dynamic DETR \cite{dai2021dynamic}, Anchor DETR \cite{wang2022anchor}, and DAB DETR \cite{liu2022dabdetr}. DAB DETR introduces a novel query formulation using anchor boxes \textit{(4D box coordinates: x, y, w, h)} in DETR and updates them layer-by-layer. This novel approach enhances spatial priors in the cross-attention module by factoring in both position and size, leading to a more straightforward implementation and a more profound insight into the role of queries in DETR. We utilize this 4D query formulation in our fine-grained detection level to initialize the decoder queries with the coarse detection results, which effectively reduces the search space for detecting drones. 

\section{METHODOLOGY}

Commensurate with real-world settings, benchmark drone-to-drone detection datasets (e.g. FL-Drones \cite{rozantsev2016detecting}, NPS-Drones \cite{li2016multi}, and AOT \cite{AOT}) contain tiny drones, occupying a mere fraction of the frame. On average, the drone size is between 0.05\% to 0.08\% of the entire frame size in these datasets. These drones exhibit rapid shape changes, even between consecutive frames, and adeptly blend into complex backgrounds such as trees and clouds. In such challenging circumstances, even a human eye struggles to localize drones, often resulting in missed detections. Notably, the initial annotations provided in \cite{rozantsev2016detecting} and \cite{li2016multi} were found to be lacking in precision, leading to revised versions released by the authors of \cite{ashraf2021dogfight}. In such scenarios, having prior information about the likely locations and sizes of drones within a frame can be immensely beneficial. By focusing our attention on these predefined regions, rather than analyzing the entire frame exhaustively, we can simplify the localization task and achieve more accurate and efficient detection results. Our Coarse-to-Fine detection approach is inspired by this intuition. 

\begin{figure*}[!]
    \centering
    \includegraphics[width=0.95\textwidth]{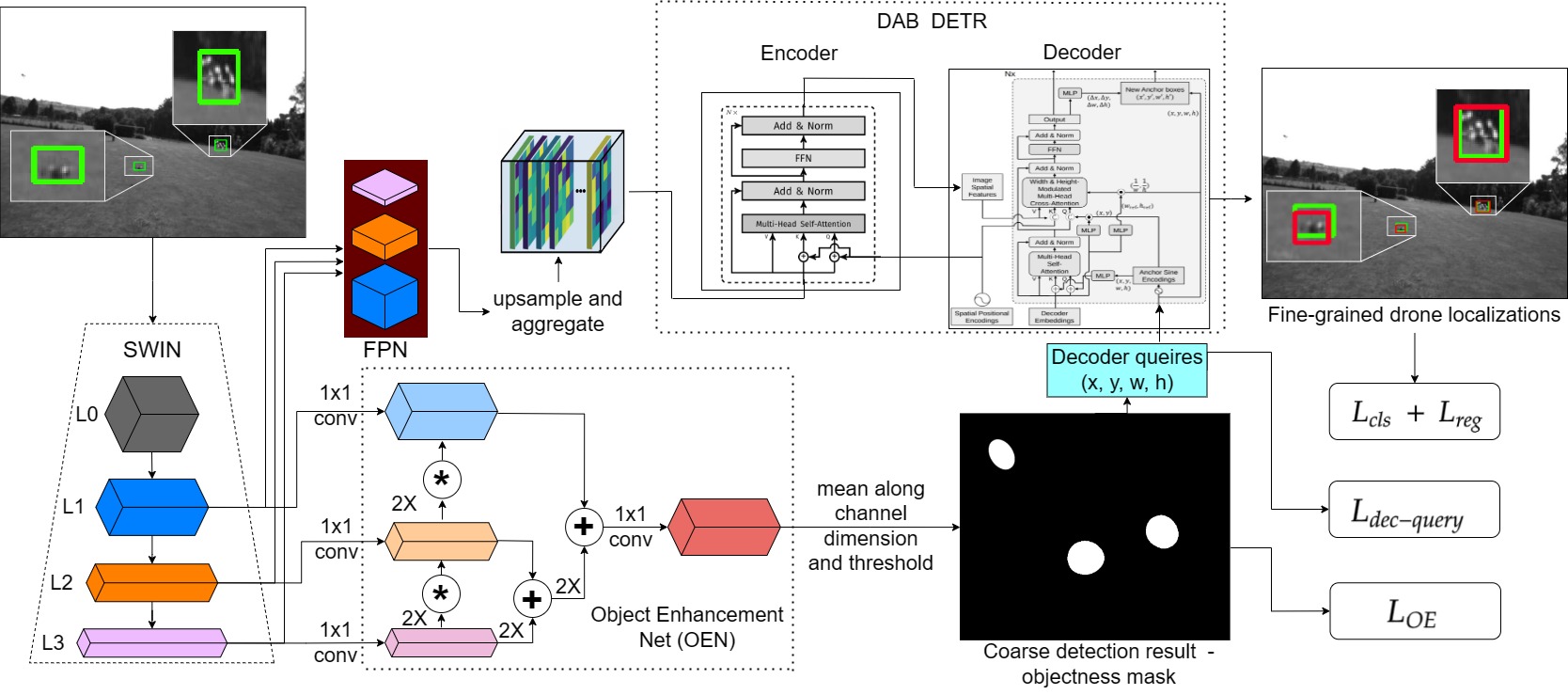}
    \caption{\textbf{Our Coarse-to-Fine detection approach}. We process video frames with the Swin Transformer \cite{liu2021swin} followed by FPN \cite{lin2017feature} to obtain multi-scale features which are input to DAB DETR \cite{liu2022dabdetr}. OEN refines the features from Swin layers 1, 2, and 3 by enhancing foreground details and reducing background noise. By computing the mean of the enhanced feature map and applying a threshold, we obtain coarse detection results that highlight the regions likely to contain objects. We utilize these regions to prime the DAB DETR decoder, significantly reducing search space and improving the localization performance. \textcolor{Green}{Green} boxes - ground truth, \textcolor{red}{Red} boxes - model predictions. $L_{cls}$ \& $L_{reg}$ are the classification and regression losses respectively, commonly used with DETR-family models \cite{liu2022dabdetr}}
    \label{fig:archi}
\end{figure*}

\vspace{0.5mm}
An overview of our proposed approach is illustrated in Fig. \ref{fig:archi}. To effectively distinguish between foreground and background elements, we employ the robust Swin Transformer \cite{liu2021swin} as our backbone architecture, which is known for capturing intricate spatial relationships and global context. To reduce the noise in the multi-scale features and amplify the inherent objectness information within them, we introduce a network called Object Enhancement Net (OEN). It takes Swin features as input, enhances foreground details, reduces background noise, and generates an objectness mask that highlights the foreground pixels. This mask serves as the initial coarse detection result. We then leverage the capabilities of the Detection Transformer (DETR) \cite{liu2022dabdetr}, recognized as the current state-of-the-art solution in detection tasks, to achieve fine-grained drone detection. More precisely, we initialize the decoder of DAB DETR \cite{liu2022dabdetr} with the coarse detection results, which represent the probable drone locations within the image. This priming process enables DETR to concentrate its attention on these regions, rather than searching for drones in the complete frame, resulting in improved drone localizations. 

\vspace{0.5mm}
In the following, we begin by delving into the specifics of the backbone network, followed by an explanation of the Object Enhancement Net (Section \ref{coarse}). Then, we elucidate the process by which we initialize the decoder of DAB DETR with the coarse detection results, yielding the drone localizations (Section \ref{fine-grained}) and lastly, the losses utilized in the training process (Section \ref{sec:losses}).

\subsection{Coarse Level: Objectness Mask} \label{coarse}

% Convolutional Neural Network (CNN) backbones employ a repetitive downsampling process using operations like max-pooling on the input image, which reduces the spatial resolution of feature maps, potentially making them less effective at detecting fine-grained local details. In contrast, attention-based backbones like Swin Transformer \cite{liu2021swin} utilize patch merging to generate multi-scale feature maps, preserving essential local details. Moreover, Swin's proficiency in capturing global context and spatial relationships enhances its ability to differentiate between foreground and background elements. Thus, we pass the input images through the Swin Transformer and obtain the spatially attended features. We pass these multi-level features (Level 2, 3, 4) through a Feature Pyramid Network (FPN) \cite{lin2017feature}, which is generally used in the object detection pipeline for detecting objects at multiple scales. 
\noindent \textit{Spatial feature extractor: }
CNN backbones employ repeated downsampling, like max-pooling, which reduces feature map resolution, potentially losing local details. In contrast, Swin Transformer \cite{liu2021swin}, an attention-based backbone, uses patch merging techniques to create multi-scale feature maps. This approach helps Swin Transformer \cite{liu2021swin} retain fine details. Additionally, it excels at capturing strong global contextual information, making it robust at identifying and localizing objects.
% , even when they are small or partially occluded. 
% which can help it differentiate between foreground and background, even when they share similar colors or textures. 
Thus, we use Swin Transformer \cite{liu2021swin} to obtain spatially attended features from input frames and pass them through a Feature Pyramid Network (FPN) \cite{lin2017feature}, which is typically used in an object detection pipeline to detect objects at multiple scales. We resize the multi-scale features generated by the FPN \cite{lin2017feature} and combine them to obtain a single feature map.

\vspace{0.8mm}
\noindent \textit{Object Enhancement Net (OEN): } \label{sec:OEN}
% ** Add the OEN diagram if space permits**
As discussed in Section \ref{sec:introduction}, due to the rapid movement of both the source and target drones in real-world drone-to-drone detection scenarios, the captured video frames often contain distortion and noise. This noise is further amplified within the feature space by downsampling operations in the backbone networks, posing a challenge in accurately detecting drones. To address this issue, drawing inspiration from \cite{Wu_2019_CVPR}, we introduce a network called Object Enhancement Net (OEN).  
% upsampling, convolutions, and element-wise multiplication operations to mitigate the noise and enhance the foreground pixels in the feature space. This network combines upsampling for spatial detail, convolutional layers for feature emphasis, skip connections for high-level context retention, and concatenation for multi-scale information integration. We refer to this module as Object Enhancement Net (OEN). 
This network reduces noise by combining upsampling for spatial detail, convolutional layers for feature emphasis, skip connections for high-level context retention, and concatenation for multi-scale information integration. It intelligently aggregates data from various Swin Transformer layers, refining feature maps to better distinguish foreground from background. Specifically, OEN takes the features from the last three layers of the Swin Transformer as input and generates a single enhanced feature map with reduced noise, which we call an Object-Enhanced (OE) feature map. %We empirically observed that this holds for attention-based backbones like Swin Transformer too. 

The mean of feature maps along the channel dimension reflects the average activation across all channels. In object detection, foreground objects typically have higher activation levels. Thus, an elevated mean along the channel dimension is a useful indicator of the presence of an object. We utilize this inherent objectness information from the image representations to generate our coarse detection results (Fig. \ref{figure_1}c). Specifically, we derive an objectness mask by averaging the OEN's output feature map along the channel dimension and applying a threshold. In our experiments, we found that a threshold of 0.6 yielded the best results. We create ground truth masks from dataset annotations, marking drone locations in a frame as white (255) and the rest as black (0). We align the objectness masks generated by OEN with the ground truth masks by employing an object enhancement loss ($\mathit{L}_{\text{OE}}$) which is a linear combination of dice loss ($\mathit{L}_{Dice}$) and instance-aware binary cross-entropy loss ($\mathit{L}_{BCE}$). Let $P$ be the objectness mask and $G$ be the ground truth mask.
% Our empirical analysis revealed that when drones are clearly visible in the frames, the mean of the feature map, obtained from the multi-scale feature fusion (output of FPN in Fig. \ref{fig:archi}), effectively highlights the drone locations. However, it fails to do so and generates a lot of noise when drones are small, blurry, or occluded. Our Object Enhancement Net (OEN) mitigates this issue. In Section \ref{sec:qualitative_res}, we present a qualitative result demonstrating the difference between the mean of the FPN feature map and the object-enhanced feature map.   %Should we point out that this a limitation in TransVisDrone?
% To address this, we use Object Enhancement Net (OEN) and accompanying Object Enhancement Losses, to amplify foreground pixels and minimize background noise in the feature maps. 

\vspace{-2mm}
% \begin{align}
%     L_{dice} &= 1 - 2\cdot\frac{|P \cap G|}{|P \cup G|}  \\
%     L_{IA-BCE} &= \sum_{i=1}^n -(G_i \cdot\log(P_i) + (1-G_i)\cdot\log(1-P_i)) \\
%     L_{obj-enhance} &= \alpha \cdot L_{dice} + \beta \cdot L_{IA-BCE} 
% \end{align}

\begin{align}
\mathit{L}_{\text{Dice}} &= 1 - 2\cdot\frac{|P \cap G|}{|P \cup G|}  \\
\mathit{L}_{\text{BCE}} &= \sum_{i=1}^n -(G_i \cdot\log(P_i) + (1-G_i)\cdot\log(1-P_i)) \\
\mathit{L}_{\text{OE}} &= \alpha \cdot \mathit{L}_{\text{Dice}} + \beta \cdot \mathit{L}_{\text{BCE}} 
\end{align}

where $n$ denotes the number of drones in a frame and $\alpha, \beta$ are hyperparameters. Via a random hyperparameter search in the range $(1,5)$ on the validation set, we found that $\alpha = 2$ and $\beta = 1$ worked best.

\subsection{Fine-grained Level: Drone Localization} \label{fine-grained}

% \noindent \textit{Detection Transformer - A primer:} DETR \cite{carion2020end} is an end-to-end trainable object detection system that eliminates the need for hand-designed components, like non-maximum suppression and anchor generation. It directly predicts object class labels and bounding box coordinates for all objects in one pass, using a Transformer-based architecture with positional encoding. The main components of a DETR include the positional encoding, the Encoder, and the Decoder. The Encoder processes image features with self-attention mechanisms to capture contextual information, while the Decoder uses queries to attend to encoded features and predict object class labels and bounding box coordinates. Numerous variants of DETR were proposed to address the slow-convergence problem of DETR, like the Deformable-DETR \cite{zhu2020deformable}, Dynamic DETR \cite{dai2021dynamic}, Anchor DETR \cite{wang2022anchor}, and DAB DETR \cite{liu2022dabdetr}. The DETR family is considered state-of-the-art in object detection due to its end-to-end training, Transformer architecture, and strong performance on benchmarks, simplifying the detection process and achieving competitive results.

To effectively utilize the coarse detection results as prior information and improve the detection performance, we utilize the strengths of DAB DETR \cite{liu2022dabdetr}. We particularly emphasize the decoder queries, which can be thought of as specific image regions that the model closely examines to identify objects and their positions.  DAB DETR \cite{liu2022dabdetr} is the first method to introduce 4D decoder queries \textit{(x, y, w, h)} and updates them layer-by-layer. This gives us more control over the position and the size of the queries. 
% Initially, in the first decoder layer, 'N' queries are randomly initialized to cover the entire image. As the encoded features progress through subsequent layers, these queries adapt to foreground objects. However, when drones are tiny and camouflaged, we observed that the queries end up missing a few drone regions. To address this, we leverage the coarse detection results obtained from the objectness mask. 
We initialize the decoder queries using the highlighted locations within the objectness mask. This strategic initialization approach effectively reduces the search space for the decoder, focusing its attention on the probable drone locations instead of scanning the entire image. To exploit temporal information from the video frames, we utilize the highlighted regions from all the frames in a batch to initialize decoder queries for each frame. Furthermore, to refine localization, we employ a loss function pushing the queries in the final decoder layer to align with the coarse detection results. Let $\mathrm{Q}$ and $\mathrm{C}$ denote the set of decoder queries (anchor boxes) and the set of highlighted regions in the objectness mask respectively. We define the decoder query loss as follows
% \begin{equation}
%     L_{dec-query} = \sum_{i=1}^{|q|} min-dist(q_i, c)
%     \vspace{-5pt}
% \end{equation}

\vspace{-1.8mm}
\begin{equation}
    \label{loss:dec_query}
    \mathit{L}_{\text{dec-query}} = \sum_{q \in \mathrm{Q}} \min_{c \in \mathrm{C}} (\text{dist} (q, c))
\end{equation}

where \textit{dist} is the Euclidean distance between a query and a highlighted region. In our experiments, we set the number of queries per image to $100$. 

\vspace{0.5mm}
Additionally, to deal with the extremely small size of drones, we restrict the size of the decoder query boxes to be less than a given constant, $A_{\max}$, thanks to the novel 4D formulation \textit{(x, y, w, h)} of decoder queries by DAB DETR \cite{liu2022dabdetr}. We define $\mathit{L}_{\text{dec-query-size}}$ as,

\vspace{-1.8mm}
\begin{equation}
    \label{loss:dec_query_size}
    \mathit{L}_{\text{dec-query-size}} = \sum_{q \in \mathrm{Q}} \max(0, |A_{q} - A_{\max}|)
\end{equation}

where $A_{q}$ is the area ($w \times h$) of decoder query $q$. We set $A_{\max}$ to $20\%$ of the frame size for NPS-Drones \cite{li2016multi} and AOT \cite{AOT} datasets, and $40\%$ for FL-Drones \cite{rozantsev2016detecting} dataset.

\subsection{Loss Functions} \label{sec:losses}

To tackle the severe foreground-to-background class imbalance, we use sigmoid focal loss \cite{lin2017focal}. For bounding box regression, we utilize both L-1 loss and GIoU loss \cite{Rezatofighi_2018_CVPR}. While IoU loss is zero when there's no overlap between ground truth and predicted bounding boxes, GIoU loss considers both overlap and spatial alignment. It penalizes predictions that have a large deviation from the ground truth in terms of both size and position, guiding the model to tightly enclose the object of interest with more accurate bounding boxes. 

\begin{table*}[!htb]

\input{tables/main_table_FL_and_NPS}
   \vspace{-3mm}
   \caption{\textbf{Detection results} comparison on FL \cite{rozantsev2016detecting} and NPS-Drones \cite{li2016multi} datasets. Values in \textbf{bold} \& \underline{underline} indicate best and second best. Percentage improvement over the State-of-the-Art, TransVisDrone, is shown inside brackets.}
   \label{tab:fl-and-nps_sota}
   \vspace{-2mm}
\end{table*}

 \begin{table}
 \footnotesize
   \input{tables/main_table_AOT}
   \vspace{-3mm}
  \caption{\textbf{Detection results} on AOT \cite{AOT} dataset.
  %Values in \textbf{bold} \& \underline{underline} indicate best and second best. Percentage improvement over the State-of-the-Art is shown inside brackets.
  (same notations as in Table \ref{tab:fl-and-nps_sota})
  }
  \vspace{-5mm}
   \label{tab:aot-sota}
\end{table}

\section{EXPERIMENTS AND RESULTS} \label{exps_section}

\subsection{Datasets}

We report our results on three challenging real-world drone-to-drone detection datasets namely FL-Drones \cite{rozantsev2016detecting}, NPS-Drones \cite{li2016multi}, and Airborne Object Tracking (AOT) dataset \cite{AOT}. For both FL-Drones and NPS-Drones datasets, we use the refined annotations provided by DogFight \cite{ashraf2021dogfight}.

\vspace{0.5mm}
\noindent \textbf{FL-Drones dataset \cite{rozantsev2016detecting}:} This dataset, although smaller in scale, presents significant challenges. The fast and erratic movements of the drones result in frequent changes in their shapes, even between consecutive frames. Moreover, the dataset exhibits substantial variations in illumination levels and minimal contrast between the drones and the background, rendering drone localization exceptionally difficult in such scenarios. Additionally, the dataset features a wide range of drone sizes, spanning from as small as 9x9 to as large as 259x197 pixels. In total, the dataset comprises 14 videos, amounting to 38,948 frames, encompassing a mixture of resolutions, including 640x480 and 752x480. In line with previous research efforts, our approach involves dividing each video into two equal parts: one for training and the other for testing.

\vspace{0.5mm}
\noindent \textbf{NPS-Drones dataset \cite{li2016multi}:} This dataset comprises high-definition (HD) images with resolutions of 1920x1280 and 1280x760 pixels. The drone sizes range from 10x8 to 65x21 pixels. A notable characteristic of this dataset is the prevalence of very small-sized drones. It encompasses 50 videos, totaling 70,250 frames. Following prior works, we use the same splits for training: videos 01-36, validation: videos 37-40, and testing: videos 41-50.  

\vspace{0.5mm}
\noindent \textbf{Airborne Object Tracking (AOT) dataset \cite{AOT}:} This dataset is hosted by Amazon Prime Air for a workshop challenge in ICCV 2021 \cite{AOT_ICCV21_challenge}. It has 5.9M+ images at the resolution of 2448 × 2048 in grayscale and 3.3M+ 2D annotations of multiple planned and unplanned airborne objects like airplanes, helicopters, birds, drones, hot air balloons, and others. The trajectories are planned to create a wide distribution of distances, closing velocities, and approach angles. For a fair comparison with the existing work \cite{sangam2023transvisdrone}, we also use part 1 of the dataset which contains a total of 987 videos split into 516 for training, 171 for testing, and 300 for validation. 
% However, we obtained better performance than (cite visdrone) by training on 250/516 training videos itself.      

\subsection{Implementation Details}

Following the prior works \cite{ashraf2021dogfight, sangam2023transvisdrone}, we train on the frames containing drones and evaluate on every 4th frame. To enhance training data diversity, we apply standard augmentations, including random horizontal flips and color jitter, with a probability of 0.5. We resize the frames to a resolution of 1920x1280. For optimization, we employ the AdamW optimizer with a learning rate set at 8e-5 and a weight decay of 1e-4. During training, we utilize a multi-step learning rate scheduler to fine-tune the learning process effectively. Furthermore, we harness the power of transfer learning by initializing our model's weights with publicly available pre-trained model weights from Swin-B and Deformable DETR, which were originally trained on the MSCOCO dataset. We train our model using two Nvidia RTX A6000 GPUs and report results using one GPU. 

\vspace{-1 pt}
\subsection{Evaluation metrics}
\vspace{-2pt}
We obtain the precision-recall curve using the all-point interpolation method and report the precision and recall values corresponding to the best F1 score. We set the IoU threshold between our model predictions and the ground truth at 0.5, compute the average of the precision values at 11 equally spaced recall points, and report this as AP@50. 

% \vspace{-0.5em}
% \begin{figure}[h]
%     \centering
%     \includegraphics[width=\linewidth, height=1.2in]{files/P-vs-R-comparison.pdf}
%     \vspace{-1em}
%     \caption{Comparison of Precision vs Recall curves between \textcolor{Red}{TransVisDrone} \cite{sangam2023transvisdrone} and \textcolor{Green}{Our method} on three benchmark datasets.}
%     \label{figure_1}
% \vspace{-2.0em}
% \end{figure}

\begin{figure}
     \centering
     \begin{subfigure}[b]{0.15\textwidth}
         \centering
         \includegraphics[width=\textwidth, height=1.3in]{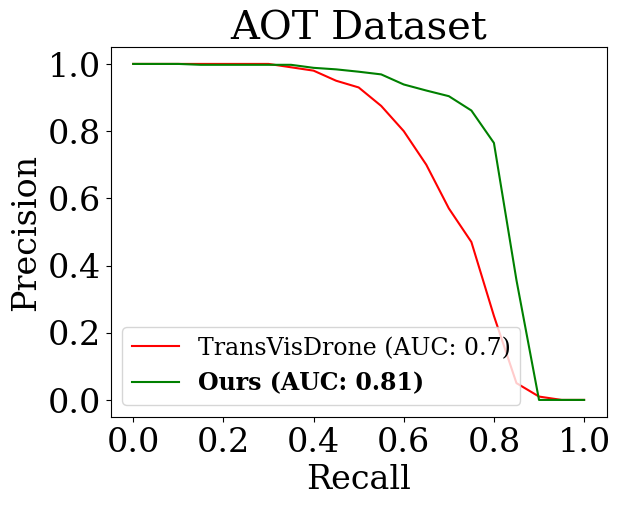}
         % \caption{$y=x$}
         \label{fig:P_R_AOT}
     \end{subfigure}
     % \hfill
     \begin{subfigure}[b]{0.15\textwidth}
         \centering
         \includegraphics[width=\textwidth, height=1.3in]{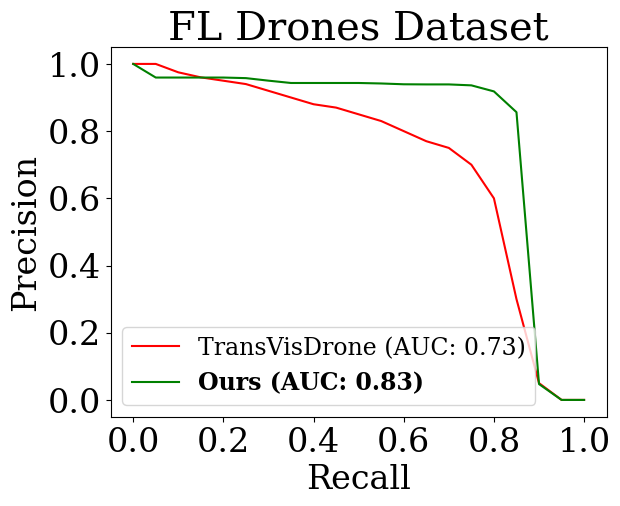}
         % \caption{$y=3\sin x$}
         \label{fig:P_R_FL}
     \end{subfigure}
     % \hfill
     \begin{subfigure}[b]{0.15\textwidth}
         \centering
         \includegraphics[width=\textwidth, height=1.3in]{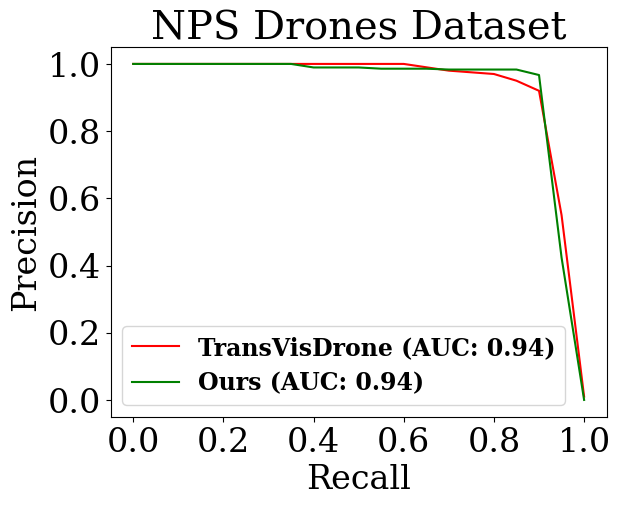}
         % \caption{$y=5/x$}
         \label{fig:P_R_NPS}
     \end{subfigure}
        \vspace{-6mm}
        \caption{\textbf{Comparison of Precision vs Recall curves} between \textcolor{Red}{TransVisDrone} \cite{sangam2023transvisdrone} and \textcolor{Green}{Our method}.}
        \label{fig:P_R_curves}
        \vspace{-16pt}
\end{figure}

\subsection{Comparison with Existing Works} \label{results}

Table \ref{tab:fl-and-nps_sota} shows the performance comparison of our coarse-to-fine detection approach with several recent methods on FL and NPS-Drones datasets. On FL-Drones, our approach achieves a significant improvement of \textbf{5\%} precision, \textbf{9\%} recall, \textbf{7\%} F1 score, and \textbf{9\%} AP when compared to the current state-of-the-art method \cite{sangam2023transvisdrone}. The FL-Drones dataset contains comparatively low-resolution frames with high distortion and noise due to the rapid motion of drones. Our model's substantial improvement in detection performance underscores the insufficiency of a simple multi-resolution feature fusion approach and highlights the superiority of a coarse-to-fine detection strategy in such challenging scenarios. On the NPS-Drones dataset, our method surpasses \cite{sangam2023transvisdrone} by \textbf{2\%} precision, \textbf{1\%} recall, and \textbf{1\%} F1 score with a comparable AP. 

\vspace{0.5mm}
Having established that our method outperforms all the existing methods on FL and NPS drone datasets, we now present its results on the AOT dataset by comparing w.r.t. the two most recent D2D detection methods, in Table \ref{tab:aot-sota}. Our approach outperforms the prior method \cite{sangam2023transvisdrone} across all the metrics with a margin of $2$-$4$\%. The outcome obtained on the AOT dataset, comprising a substantial 5.9 million high-resolution images, clearly highlights the effectiveness of our approach, emphasizing its adaptability and practical utility in real-world scenarios.
% This result on the AOT dataset, which contains a staggering $5.9$ million high-resolution images, unequivocally underscores the potency of our approach underscoring its scalability and real-world applicability.

\subsection{Ablation Studies} \label{sec:ablation_studies}
% \vspace{-2mm}

% \vspace{1em}
In this section, we present results that validate the effectiveness of various components in our proposed approach. Using Swin \cite{liu2021swin} + DAB DETR \cite{liu2022dabdetr} as the baseline, Table \ref{tab:ablations} showcases the enhancements in detection performance achieved by incorporating each component of our approach.

\begin{table}[h]
 \footnotesize
   \input{tables/ablations}
   \vspace{-3mm}
  \caption{\textbf{Ablation:} Study of different components of our method on FL-Drones dataset (@640 resolution)}
  \vspace{-5mm}
   \label{tab:ablations}
\end{table}

 \begin{table}[h]
 \footnotesize
   \input{tables/spatial_resolution}
   \vspace{-3mm}
  \caption{\textbf{Sensitivity to Image Resolution:} Study on the effect of spatial resolutions on our model's performance on the FL-Drones dataset.}
  \vspace{-5mm}
   \label{tab:spatial-res}
\end{table}

 \begin{table}[h]
 \footnotesize
   \input{tables/backbones}
   \vspace{-3mm}
  \caption{\textbf{Sensitivity to Backbones:} Study on different backbones on FL-Drones dataset (@640 resolution). %(Note: Swin-S provided about 43 FPS, while Swin-T went upto 58 FPS in our studies.)
  }
  % \vspace{-2mm}
   \label{tab:backbones}
\end{table}

% \vspace{em}
We also investigate the impact of spatial resolutions of frames and different backbones while providing insights into the trade-off between performance and throughput in Tables \ref{tab:spatial-res} and \ref{tab:backbones}. Notably, our top-performing model, utilizing the Swin-B backbone, achieves an impressive FPS rate exceeding 35 and the model that achieves the highest throughput (Swin-T backbone) outperforms the prior work \cite{sangam2023transvisdrone} by a significant margin.

\subsection{Qualitative Results} \label{sec:qualitative_res}

\begin{figure}[!t] 
    \centering
    \includegraphics[width=0.95\linewidth, height=2in]{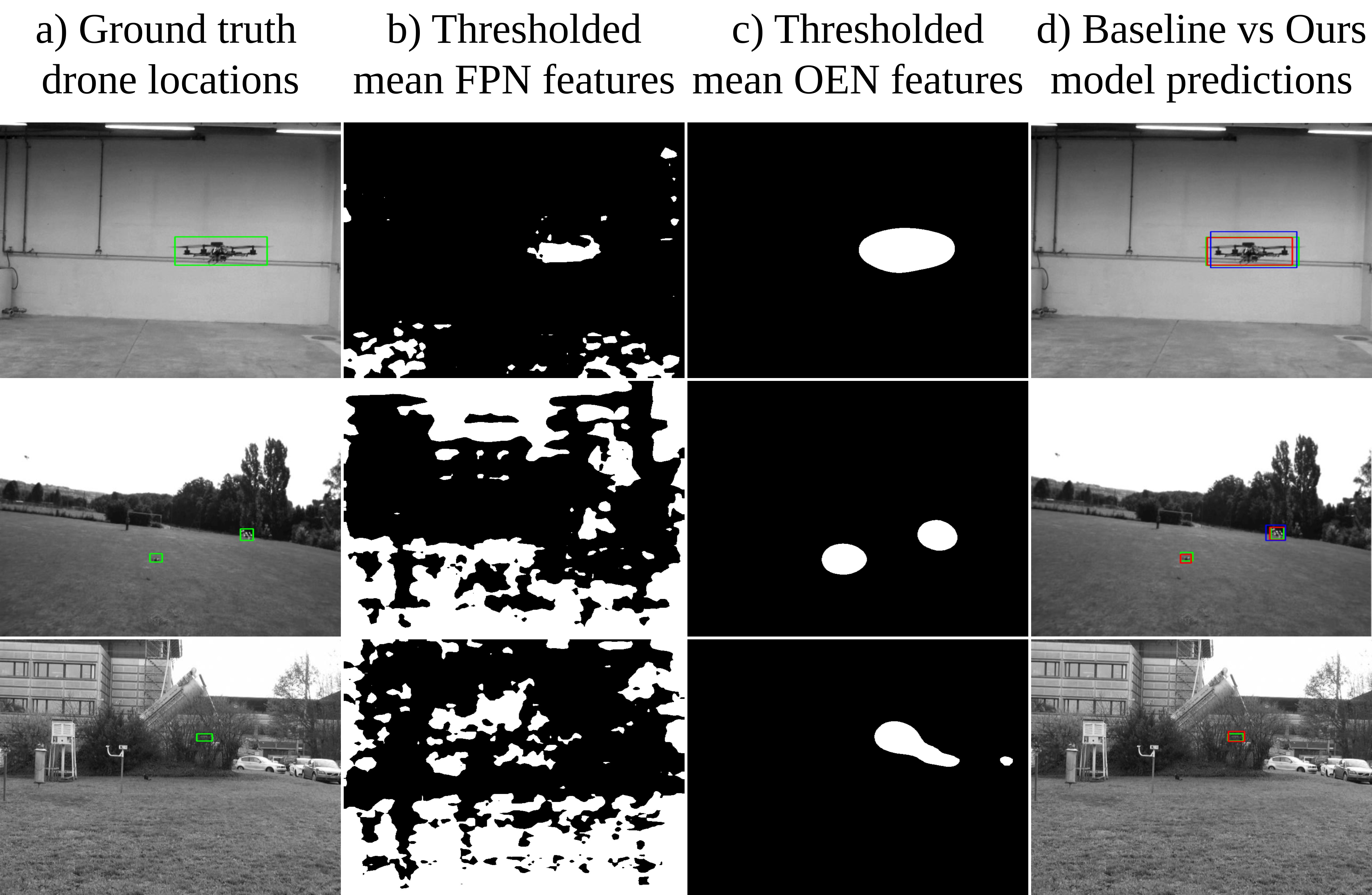}
    % \vspace{-1mm}
    \caption{
    % \textbf{Coarse and Fine-grained results: Baseline vs Our model.} We consider Swin \cite{liu2021swin} $+$ FPN \cite{lin2017feature} $+$ DAB DETR \cite{liu2022dabdetr} as the baseline. \textcolor{Green}{Green} - ground truth, \textcolor{Blue}{Blue} - baseline predictions and \textcolor{Red}{Red} box - our model predictions.
    \textbf{Qualitative Analysis:} We use coarse-level localization information of drones to guide the DAB-DETR decoder queries (Equation \ref{loss:dec_query}). Traditional FPN features contain severe noise (Column b), which is mitigated by our proposed Object Enhancement Network (Column c), leading to accurate drone detections in challenging scenarios (Column d). \textcolor{Green}{Green} - ground truth, \textcolor{Blue}{Blue} - baseline predictions and \textcolor{Red}{Red} box - our model predictions.
    % We consider Swin \cite{liu2021swin} $+$ FPN \cite{lin2017feature} $+$ DAB DETR \cite{liu2022dabdetr} as the baseline. \textcolor{Green}{Green} - ground truth, \textcolor{Blue}{Blue} - baseline predictions and \textcolor{Red}{Red} box - our model predictions.
    }
    \label{fig:qual_res}
    \vspace{-6mm}
\end{figure}

Figure \ref{fig:qual_res} presents a comparative analysis between the outcomes achieved by the baseline model and our coarse-to-fine detection approach when applied to a selection of challenging frames from the FL-Drones dataset \cite{rozantsev2016detecting}. The second column of the figure demonstrates how the downsampling operations utilized within the backbone networks tend to exacerbate the noise within the feature space. Notably, our proposed Object Enhancement Network (OEN) module effectively mitigates this noise, consequently accentuating the foreground pixels, as depicted in the third column. This enhancement plays a pivotal role in our model's ability to achieve precise drone localization at the fine-grained level when compared to the baseline. In particular, column 4 (rows 2 and 3) highlights instances where the baseline model struggles to detect drones that are either minuscule or seamlessly blend into the background, whereas our model adeptly identifies and localizes them with accuracy.

\

% \vspace{-0.3em}
% \begin{figure*}[!t]
%     \centering
%     \includegraphics[width=0.95\textwidth, height=3in]{files/Qual_Res.pdf}
%     \vspace{-2mm}
%      \caption{\textbf{Coarse and Fine-grained results: Baseline vs Our model.} We consider Swin \cite{liu2021swin} $+$ FPN \cite{lin2017feature} $+$ DAB DETR \cite{liu2022dabdetr} as the baseline. \textcolor{Green}{Green} - ground truth, \textcolor{Blue}{Blue} - baseline predictions and \textcolor{Red}{Red} box - our model predictions.}
%     \label{fig:qual_res}
%     \vspace{-4mm}
% \end{figure*} 

\vspace{-6mm}
\section{Drone-to-Drone detection in real world}
% \vspace{-2pt}
% \subsection{Deployment on Edge Devices}
\subsection{Edge computing deployment}
% \vspace{-2pt}
% To verify the real-time processing capability of our model, we deployed it on an NVIDIA Jetson Xavier NX \cite{jetson} board. 
To validate the real-world applicability of our model, we deployed it on an NVIDIA Jetson Xavier NX \cite{jetson} board. Our model using the Swin-T backbone achieved a real-time performance of 31 FPS on 640-resolution frames.  
% \vspace{-3pt}
\subsection{Minimal False Positives}
% \vspace{-5pt}
Low false positives in real-time drone-to-drone detection systems are essential for safety, efficiency, and trust, as they prevent unnecessary disruptions, conserve resources, and ensure compliance with regulations. To validate the effectiveness of our approach, we assessed False Positives Per Image (FPPI) using the AOT dataset's 194,193 test frames. Our method achieved an impressively low FPPI of 
\textbf{3.2e-4}, vs 4.4e-4 (TransVisDrone \cite{sangam2023transvisdrone}), 1.8e-2 (DogFight \cite{ashraf2021dogfight}) and 2.5e-2 (De-DETR \cite{zhu2020deformable}), highlighting its precision.
% \textbf{0.000324}, vs 0.000437 (TransVisDrone \cite{sangam2023transvisdrone})
% , 0.018 (DogFight \cite{ashraf2021dogfight}) and 0.02474 (De-DETR \cite{zhu2020deformable}), highlighting its precision.

% \vspace{-5pt}
\section{CONCLUSIONS}
% \vspace{-3pt}
We have introduced a cost-effective vision-based system for drone-to-drone detection. Unlike existing methods, we propose a coarse-to-fine detection strategy leveraging the vision transformer networks and harnessing the untapped objectness information present in the image representations. Our model is designed to be end-to-end trainable and achieves real-time performance. We will make our code base publicly available. 

\addtolength{\textheight}{-1cm}   % This command serves to balance the column lengths

%%%%%%%%%%%%%%%%%%%%%%%%%%%%%%%%%%%%%%%%%%%%%%%%%%%%%%%%%%%%%%%%%%%%%%%%%%%%%%%%

%%%%%%%%%%%%%%%%%%%%%%%%%%%%%%%%%%%%%%%%%%%%%%%%%%%%%%%%%%%%%%%%%%%%%%%%%%%%%%%%

%%%%%%%%%%%%%%%%%%%%%%%%%%%%%%%%%%%%%%%%%%%%%%%%%%%%%%%%%%%%%%%%%%%%%%%%%%%%%%%%
% \vspace{-5pt}
\section*{ACKNOWLEDGMENTS}

We are grateful to the Ministry of Electronics and Information Technology and Ministry of Education, Govt of India, as well as IIT-Hyderabad through its MoE-DRDO fellowship program for their support of this project. We thank Joseph KJ for his insightful discussions. We also express our gratitude to Charchit Sharma for his valuable assistance in configuring the AOT dataset.

\bibliographystyle{IEEEtran}
\bibliography{IEEEabrv,references}

\end{document}

%% file: tables/main_table_FL_and_NPS.tex
\begin{center}
    \scalebox{0.9}{
    \begin{tabular}{cc|c|c|c|c|c|c|c|c}
        \toprule
        \multirow{2}{*}{Method} & \multirow{2}{*}{Venue} & \multicolumn{4}{c|}{FL-Drones} & \multicolumn{4}{c}{NPS-Drones} \\
        \cmidrule(lr){3-6} \cmidrule(lr){7-10} % Use \cmidrule instead of \cline for better spacing
        & &  Precision & Recall & F1 Score & AP@50 & Precision & Recall & F1 Score & AP@50 \\
        \midrule
        % Add your data rows here
        Mask-RCNN~\cite{He_2017_ICCV} & ICCV'17  & 0.76 & 0.68 & 0.72 & 0.68 & 0.66 & 0.91 & 0.76 & 0.89\\ 
        SRCDet-H~\cite{yang2019scrdet}& ICCV'19  & 0.54 & 0.62 & 0.58 & 0.52 & 0.81 & 0.74 & 0.77 & 0.65\\
        SRCDet-R~\cite{yang2019scrdet}& ICCV'19  & 0.55 & 0.62 & 0.58 & 0.52 & 0.79 & 0.71 & 0.75 & 0.61\\
        SLSA ~\cite{wu2019sequence} & ICCV'19    & 0.57 & 0.72 & 0.64 & 0.61 & 0.47 & 0.67 & 0.55 & 0.46\\
        FCOS~\cite{tian2019fcos} & ICCV'19       & 0.69 & 0.70 & 0.69 & 0.62 & 0.88 & 0.84 & 0.86 & 0.83\\
        MEGA~\cite{chen2020memory} & CVPR'20     & 0.71 & 0.72 & 0.71 & 0.65 & 0.88 & 0.82 & 0.85 & 0.83\\
        
        DogFight~\cite{ashraf2021dogfight} & CVPR'20 & 0.84 & 0.76 & 0.80 & 0.72 & 0.92 & 0.91 & 0.92 & 0.89\\
        De-DETR~\cite{zhu2020deformable} & ICLR'21 & 0.72 & 0.70 & 0.71 & 0.64 & 0.85 & 0.80 & 0.82 & 0.76\\
        % TransVisDrone~\cite{sangam2023transvisdrone} & ICRA'23 & \textcolor{blue}{0.84} & \textcolor{blue}{0.76} & \textcolor{blue}{0.80} & \textcolor{blue}{0.75} & \textcolor{blue}{0.92} & \textcolor{blue}{0.91} & \textcolor{blue}{0.92} & \textcolor{red}{0.95  }\\
        TransVisDrone~\cite{sangam2023transvisdrone} & ICRA'23 & \underline{0.84} & \underline{0.76} & \underline{0.80} & \underline{0.75} & \underline{0.92} & \underline{0.91} & \underline{0.92} & \textbf{0.95}\\
        \midrule
        % Ours &  & \textcolor{red}{0.89 ($+5\%$)}  & \textcolor{red}{0.85 ($+9\%$)} & \textcolor{red}{0.87 ($+7\%$)} & \textcolor{red}{0.84  ($+9\%$)}  & \textcolor{red}{0.94 ($+2\%$)} & \textcolor{red}{0.92 ($+1\%$)} & \textcolor{red}{0.93 ($+1\%$)} & \textcolor{blue}{0.93}\\ 
        \rowcolor{Gray}
        Ours &  & \textbf{0.89 \textcolor{blue}{($\mathbf{+5\%}$)}}  & \textbf{0.85 \textcolor{blue}{($\mathbf{+9\%}$)}} & \textbf{0.87 \textcolor{blue}{($\mathbf{+7\%}$)}} & \textbf{0.84  \textcolor{blue}{($\mathbf{+9\%}$)}}  & \textbf{0.94 \textcolor{blue}{($\mathbf{+2\%}$)}} & \textbf{0.92 \textcolor{blue}{($\mathbf{+1\%}$)}} & \textbf{0.93 \textcolor{blue}{($\mathbf{+1\%}$)}} & \underline{0.93}\\ 
        \bottomrule
    \end{tabular}
}
\end{center}

%% file: tables/main_table_AOT.tex
\begin{center}
\scalebox{0.85}{
    \begin{tabular}{c|c|c|c|c}
        \toprule
        Method & Precision & Recall & F1 score & AP@50 \\
        \midrule
        DogFight~\cite{ashraf2021dogfight} & 0.82 & 0.65 & 0.73 & 0.74 \\
        TransVisDrone~\cite{sangam2023transvisdrone} & \underline{0.82} & \underline{0.72} & \underline{0.77} & \underline{0.80} \\
        \midrule
        \rowcolor{Gray}
        Ours & \textbf{0.85 \textcolor{blue}{($\mathbf{+3\%}$)}} & \textbf{0.76 \textcolor{blue}{($\mathbf{+4\%}$)}} & \textbf{0.80 \textcolor{blue}{($\mathbf{+3\%}$)}} & \textbf{0.82 \textcolor{blue}{($\mathbf{+2\%}$)}} \\ 
        \bottomrule
    \end{tabular}
    }
\end{center}

% \begin{center}
% \scalebox{0.85}{
%     \begin{tabular}{c|c|c|c|c}
%         \toprule
%         Method & Precision & Recall & F1 score & AP@50 \\
%         \midrule
%         DogFight & 0.82 & 0.65 & 0.73 & 0.74 \\
%         TransVisDrone  & \underline{0.82} & \underline{0.72} & \underline{0.77} & \underline{0.80} \\
%         \midrule
%         \rowcolor{Gray}
%         Ours & \textbf{0.85 \textcolor{blue}{($\mathbf{+3\%}$)}} & \textbf{0.76 \textcolor{blue}{($\mathbf{+4\%}$)}} & \textbf{0.80 \textcolor{blue}{($\mathbf{+3\%}$)}} & \textbf{0.82 \textcolor{blue}{($\mathbf{+2\%}$)}} \\ 
%         \bottomrule
%     \end{tabular}
%     }
% \end{center}

%% file: tables/ablations.tex
\begin{center}
\scalebox{0.9}{
    \begin{tabular}{c|c|c|c|c}
        \toprule
        Method & Precision & Recall & F1 Score & AP@50 \\
        \midrule
        Swin \cite{liu2021swin} + DAB-DETR \cite{liu2022dabdetr} & 0.78 & 0.68 & 0.73 & 0.68 \\
        + OEN with OE Losses & 0.80 & 0.70 & 0.75 & 0.69 \\
        + Initialize Decoder Queries & 0.85 & 0.74 & 0.79 & 0.72 \\
        + Decoder Query Losses & \textbf{0.89} & \textbf{0.81} & \textbf{0.85} & \textbf{0.81} \\ 
        \bottomrule
    \end{tabular}
    }
\end{center}

%% file: tables/spatial_resolution.tex
\begin{center}
    \begin{tabular}{c|c|c|c|c}
        \toprule
        Frame Resolution & Precision & Recall & F1 Score & AP@50 \\
        \midrule
        Image@640 & 0.89 & 0.81 & 0.85 & 0.81 \\
        Image@800 & 0.89 & 0.83 & 0.86 & 0.82 \\
        Image@1280 & \textbf{0.89} & \textbf{0.85} & \textbf{0.87} & \textbf{0.84} \\
        \bottomrule
    \end{tabular}
\end{center}

%% file: tables/backbones.tex
\begin{center}
    \begin{tabular}{c|c|c|c|c|c}
        \toprule
        Backbone & Precision & Recall & F1 Score & AP@50 & Real-time?\\
        &&&&&($>$ 35 fps)\\
        \midrule
        Swin-T & 0.88 & 0.79 & 0.83 & 0.79 & \checkmark \\
        Swin-S & 0.89 & 0.80 & 0.84 & 0.80 & \checkmark \\
        Swin-B & \textbf{0.89} & \textbf{0.81} & \textbf{0.85} & \textbf{0.81} &  \checkmark \\
        \bottomrule
    \end{tabular}
\end{center}